\pdfoutput=1
\documentclass[letterpaper]{article}
\usepackage{workshop}
\usepackage{times}
\usepackage{helvet}
\usepackage{courier}
\usepackage[hyphens]{url}
\usepackage{graphicx}
\usepackage{natbib}
\usepackage{caption}
\usepackage{amsmath,amssymb,amsthm}
\usepackage{booktabs}
\usepackage{multirow}
\newtheorem{proposition}{Proposition}
\newcommand{\Wone}{W_1}
\newcommand{\excess}{\Delta_{\mathrm{x}}}
\newcommand{\cvar}{\mathrm{CVaR}}
\newcommand{\Pmodel}{P_\theta}
\newcommand{\Penv}{P_{\mathrm{env}}}

\nocopyright
\title{When Can Learned Uncertainty Be Trusted?\\An Audit of Distributional Reinforcement Learning}
\author{Hari Prasad}
\affiliations{h4ri.prasad@gmail.com}

\begin{document}
\maketitle

\begin{abstract}
Many learned predictive models represent their uncertainty as a
probability distribution, and planners and policies use this
distribution to judge risk. However, the fact that a model outputs a
distribution does not tell us whether that distribution captures the
uncertainty in the environment which is relevant for a decision. In this
work we study this question in distributional reinforcement learning,
where the learned return distribution $\Pmodel(Z \mid s,a)$ can be
compared directly with an environment-grounded distribution
$\Penv(Z \mid s,a)$ estimated by snapshot-restart Monte Carlo. Our audit
first screens for the claims that a risk-sensitive user would act upon,
using the \emph{excess Wasserstein gap} between the top two actions
(which is equal to the mass by which first-order stochastic dominance is
violated). Each claim is then tested against ground truth using
permutation nulls, bootstrap refutation and FDR control. Across
QR-DQN, C51 and IQN on MinAtar (33 runs), 40--95\% of the strongest
claimed risk trade-offs are refuted at 95\% confidence, their placement
is statistically indistinguishable from a truth-blind placement, and
almost none of them can be confirmed. A similar pattern is seen for a
pretrained full-Atari QR-DQN. On positive controls of known magnitude
the same audit confirms 96--100\% of real claims (correlation
$0.89$--$0.92$), so it is able to detect real risk when it is present.
Acting on the learned CVaR advice at the most-flagged states gives
results ranging from beneficial to significantly \emph{worse than
chance}. Ensembling, risk-sensitive training and recalibration do not
fix the problem, and the recalibration experiment indicates that the
learned spread has no usable information about the true risk. Based on
these results, we suggest that uncertainty estimates, including those
of learned world models, should be checked against realized outcomes at
the decisions where they are used, in addition to aggregate measures of
distributional fit.
\end{abstract}

\section{Introduction}

Many learned predictive models give a distribution as output in place of
a point estimate. Probabilistic dynamics ensembles and latent world
models predict distributions over future states and rewards
\citep{chua2018pets,ha2018worldmodels,hafner2020dreamer}, and
distributional value functions predict distributions over returns
\citep{bellemare2017c51,dabney2018qrdqn,dabney2018iqn}. These
distributions are often used directly. For example, a planner may look at
the spread of the prediction to decide which action is safer. A
predictive distribution can however fail to represent the uncertainty of
the environment. Approximation error, bootstrapping, representation
artifacts and optimization noise can all produce a spread which looks
reasonable and is a valid probability distribution, but which has no
counterpart in the realized outcomes. The question we try to answer in
this paper is therefore \emph{when the uncertainty estimates of a
learned model can be trusted for decision-making}.

To answer this, we need a setting in which the predictive distribution
of the model can be compared with the true distribution of the
environment, state by state. Distributional reinforcement learning (RL)
is a convenient choice for this. The learned return distributions of
distributional agents are already taken at face value in practice. They
are visualized to interpret agent behavior
\citep{greydanus2018visualizing}, thresholded to obtain risk-sensitive
policies via conditional value-at-risk (CVaR) and related functionals
\citep{keramati2020cvar,lim2022risksensitive}, and monitored as safety
signals in applications ranging from autonomous driving
\citep{hoel2021eqn} to safe-RL pipelines in general
\citep{garcia2015safe}. Also, since simulators can be restarted from
snapshots, the environment-grounded return distribution can be estimated
by Monte Carlo at any state we wish to audit. So the audit compares
\[
\underbrace{\Pmodel(Z \mid s,a)}_{\text{learned uncertainty}}
\quad\text{vs.}\quad
\underbrace{\Penv^{\pi}(Z \mid s,a)}_{\text{environment uncertainty}},
\]
which is the same comparison one would ideally like to make for a world
model, namely $p_\theta(\text{future}\mid s,a)$ vs.\
$p_{\mathrm{env}}(\text{future}\mid s,a)$.

There are some reasons to doubt the learned distributions even in this
setting. Quantile temporal-difference (QTD) learning converges to fixed
points which need not be equal to the true return distribution
\citep{rowland2023qtd}. The benefits of distributional methods are seen
even when only the mean is used
\citep{rowland2023statistical,lyle2019comparative}, and the spread of a
quantile head mixes up aleatoric and epistemic uncertainty
\citep{clements2019estimating}. What is missing, in our view, is a
\emph{measurement} that is targeted at decisions. A distribution can be
wrong in ways which do not affect anybody's decision, or it can be wrong
at exactly those places where a risk-sensitive user makes use of it, and
aggregate divergence or calibration metrics cannot tell the two cases
apart. Hence, apart from asking whether the predicted distribution is
accurate, we also ask whether the risk difference it reports between two
actions is actually present in the environment.

\textbf{Contributions.}
\textbf{(1) A decision-relevant audit of learned uncertainty.} We screen
states using the \emph{excess Wasserstein gap} $\excess(s)$ between the
top two actions. This quantity is positive exactly when neither action
first-order stochastically dominates the other
(Prop.~\ref{prop:dominance}), i.e.\ when the model claims a risk
trade-off that would change a risk-sensitive decision. Each such claim
is tested against snapshot-restart ground truth with permutation nulls,
bootstrap refutation and FDR control \citep{benjamini1995fdr}.
\textbf{(2) Main finding.} Across QR-DQN, C51 and IQN on three MinAtar
games (33 runs), 40--95\% of the strongest claimed trade-offs are
refuted, their placement is indistinguishable from truth-blind, and
almost none of them are confirmable. For a pretrained full-Atari QR-DQN,
\emph{every} top Breakout claim is refuted. On positive controls of
known magnitude, 96--100\% of real claims are confirmed.
\textbf{(3) Decision-level consequences.} Acting on the learned CVaR
advice at the most-flagged states can be informative or can be
significantly \emph{worse than chance}, and we could not find any
observable signal which separates the two cases.
\textbf{(4) Structure and repairs.} The artifact is already fully formed
early in training, it is uncorrelated with score, and it differs from
seed to seed. Ensembling, risk-sensitive training and recalibration do
not give usable claims.
\textbf{(5) Two pitfalls} of restart-based evaluation (RNG cloning and
the noise floor of the empirical Wasserstein distance). Each of these
gave us a convincing but wrong audit, and both apply to any
simulator-based validation of predictive distributions.

\subsection{Connection to Learned World Models}
\label{sec:wm}

Our experiments are on distributional RL and not on learned dynamics
models. The underlying methodological question is however the same,
namely \emph{whether a learned predictive distribution gives a
trustworthy picture of uncertainty for the decisions which depend on
it}. A world model may represent many possible future trajectories,
observations or states, and a planner may use the dispersion of these
predictions to assess risk \citep{chua2018pets}. The predictive
distribution by itself does not guarantee that its uncertainty follows
the variability of outcomes in the environment. Model error,
approximation error and representation artifacts can make a model look
confident, or uncertain, in ways that do not show up as corresponding
differences in the realized outcomes.

Our audit gives one way to test this correspondence with decisions in
mind. Aggregate metrics such as likelihood, calibration and
distributional distance \citep{guo2017calibration,kuleshov2018calibrated}
look at the uncertainty as a whole. We instead identify the states where
the learned distribution implies a materially different risk between
actions, and then test whether these differences are supported by
environment-grounded samples. Return distributions make this test
simple to carry out. The return is a one-dimensional summary of the
future, it is the quantity that a risk-sensitive decision rule uses, and
its true distribution can be estimated by repeated rollouts. For these
reasons we use distributional RL as a controlled testbed for a question
that applies to learned predictive models in general, i.e.\ whether the
uncertainty reported by a model matches the uncertainty in the
environment that affects decisions. We do not claim that distributional
value heads are world models, or that our results carry over to world
models without further testing. In \S\ref{sec:discussion} we discuss
what the audit suggests for them.

\section{Related Work}
\label{sec:related}

\textbf{Distributional RL.} C51 \citep{bellemare2017c51}, QR-DQN
\citep{dabney2018qrdqn} and IQN \citep{dabney2018iqn} established the
deep distributional family \citep{bdr2023book}. It has been shown that
these methods help even without distributional accuracy
\citep{lyle2019comparative,rowland2023statistical}, and that QTD fixed
points are non-unique and need not be equal to the true return
distribution \citep{rowland2018analysis,rowland2023qtd}. Our work gives
an empirical, decision-level counterpart to these theoretical warnings.

\textbf{Uncertainty (mis)estimation.} The spread of a quantile head
mixes epistemic and aleatoric components
\citep{clements2019estimating,kendall2017uncertainties}. Later works
have proposed ensembles
\citep{hoel2021eqn,eriksson2022sentinel,charpentier2022disentangling},
projection-ensemble diversity \citep{zanger2024dpe}, evidential
calibration \citep{stutts2024ceqr} and conformal wrappers
\citep{gan2025conformal}. All these methods assume from the start that the
raw distributions are unreliable, and they build estimators around them.
To our knowledge none of them measures, state by state, whether the risk
claims of a trained model are true. This is what our audit provides.

\textbf{Evaluation methodology.} Calibration metrics
\citep{guo2017calibration,kuleshov2018calibrated} and distributional OPE
with error guarantees \citep{wu2023fle} look at accuracy in aggregate or
under known conditions. We are not aware of earlier audits of deep
agents that work at the level of individual states, control for
statistical error, and target decisions. Our harness is in line with the
push for more rigorous deep-RL evaluation
\citep{machado2018revisiting,agarwal2021precipice}. It is also similar
in spirit to \citet{atrey2020exploratory}, who showed that saliency
explanations of Atari agents do not hold up under counterfactual
testing. The noise floor of the empirical Wasserstein distance, which is
behind our second pitfall, is a classical result \citep{fournier2015rate}.

\section{Auditing Learned Uncertainty Against the Environment}
\label{sec:metric}

\textbf{Learned vs.\ environment uncertainty.} Let
$Z^\pi(s,a)=\sum_t \gamma^t R_t$ be the random discounted return
obtained by taking action $a$ in state $s$ and following policy $\pi$
thereafter. Its law $\Penv^\pi(Z\mid s,a)$ comes entirely from the
stochasticity of the environment. A distributional agent predicts
$\Pmodel(Z\mid s,a)$. QR-DQN \citep{dabney2018qrdqn} does this with $N$
quantile atoms at fixed levels, C51 \citep{bellemare2017c51} with
learned masses on $N$ fixed atom locations, and IQN with an implicit
quantile function. A risk-sensitive user makes use of $\Pmodel$ through
a functional like $\cvar_\alpha(Z) = \mathbb{E}[Z \mid Z \le
F_Z^{-1}(\alpha)]$ \citep{artzner1999coherent,rockafellar2000cvar}. Such
a functional differs from mean-greedy behavior only at states where the
distributions of the actions differ in \emph{shape}, and a shift in
location alone does not change it. We audit $\Pmodel$ in the form in
which it is used. Whether or not the head is meant to recover
$\Penv^\pi$ exactly, the interpretability, CVaR and monitoring uses
mentioned above treat it as if it does, and it is this usage that we
test.

\textbf{The excess Wasserstein gap.} For the two actions $a_1, a_2$
which are ranked top-2 by mean at $s$, the 1-Wasserstein distance
between their quantile representations is $\Wone = \frac{1}{N}\sum_i
|\theta_i(s,a_1) - \theta_i(s,a_2)|$, which can be computed in closed
form. Writing $\mu_j(s) = \mathbb{E}Z(s,a_j)$, we define
\[
\excess(s) = \Wone\big(Z(s,a_1), Z(s,a_2)\big) -
\big|\mu_1(s) - \mu_2(s)\big| \ge 0 .
\]

\begin{proposition}
\label{prop:dominance}
For real-valued distributions with CDFs $F, G$,
$\excess = \int |F - G| \,dx - \big|\int (F - G)\, dx\big| = 0$ iff $F -
G$ does not change sign, i.e.\ iff one distribution first-order
stochastically dominates the other. Otherwise $\excess$ is equal to
twice the mass of the smaller-signed region of $F - G$, which is the
magnitude of the dominance violation.
\end{proposition}

\noindent The proof follows in one line from $\Wone(F,G)=\int|F-G|$ and
$\mathbb{E}G-\mathbb{E}F=\int(F-G)$ \citep{villani2009ot}, and is given
in the supplement.

When $\excess(s) > 0$, the model is claiming that the two actions trade
off risk. Neither action is safer at every quantile, and so
risk-sensitive and mean-greedy decisions can disagree at $s$. When
$\excess(s) = 0$, every monotone risk functional ranks the two actions
in the same order. As a screening statistic, $\excess$ picks out the
uncertainty which has decision content and ignores errors that are
common to both actions. That is, we audit \emph{relative} risk
claims, since these are the ones that are acted upon. The same
construction can be applied to any model which outputs per-action
outcome distributions, including the returns imagined by a world model.

\section{Audit Methodology}
\label{sec:method}

We give a summary of the method here. Full details are in the
supplement.

\textbf{Screening and stratified sampling.} We roll out the trained
agent ($\epsilon$-greedy, $\epsilon = 0.05$) and collect $2{,}000$
unique states, score each state by $\excess$ using the learned head, and
rank them. Since overclaiming could be concentrated anywhere in the
ranking, we audit seven states from each of seven rank strata (top
$0.5\%$, $0.5$--$2\%$, $2$--$5\%$, $5$--$10\%$, $10$--$25\%$,
$25$--$50\%$, $50$--$100\%$), giving 49 states per run. By ``top
strata'' we mean the top 2\% of the ranking, which are the states a
practitioner looking at the model would flag first.

\textbf{Environment-grounded uncertainty by snapshot restart.} We store
deep copies of the environment at up to 16 visits of each state. For
each audited state and each of the top-2 actions we run $2{,}000$
rollouts. In each rollout we restart from a stored snapshot, force the
action, and then follow the greedy policy until termination or up to a
horizon of $10^3$. Every restarted copy is reseeded independently
(pitfall P1). The only source of noise is the stochasticity of the
environment (sticky actions, random spawns, difficulty ramping), so the
rollouts are samples from $\Penv^\pi$. The empirical MC excess
$\hat\excess^{\mathrm{MC}}(s)$ is computed with the same quantile
formula.

\textbf{Statistical harness.} A claim is \emph{confirmed} if
$\hat\excess^{\mathrm{MC}}(s)$ is larger than a permutation null. The
null is the excess between two split halves of the rollouts of one
action, which by construction contains only estimation noise. A claim is
\emph{refuted} if the learned $\excess(s)$ is larger than the 97.5th
bootstrap percentile of the MC excess. The finite-sample bias of the
estimator (P2) pushes this percentile upwards, so the refutation test is
conservative. Both tests use Benjamini--Hochberg
\citep{benjamini1995fdr} at FDR $0.10$, pooled per game. The median null
is $0.07$, which is below the typical top-stratum claim, so real
trade-offs above ${\approx}0.07$ can be detected. We also compute a
per-state \emph{aliasing index} (the dispersion of per-snapshot mean
returns divided by the total spread). This index flags states where the
observation may hide part of the environment state, and where some
learned uncertainty would therefore be justified. Results for low-alias
states ($<0.25$) are reported separately.

\subsection{Two Pitfalls That Produce Convincing, Wrong Audits}
\label{sec:pitfalls}

\begin{figure}[t]
\centering
\includegraphics[width=0.62\columnwidth]{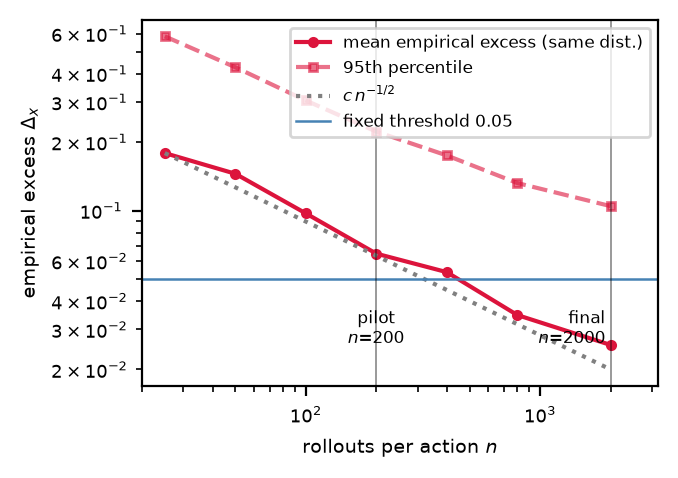}
\caption{Pitfall P2, measured. Empirical excess between two $n$-samples
from the \emph{same} return distribution (a high-variance audited
Breakout state, $\sigma \approx 2.3$), with 300 replications per $n$.
The floor decays as $n^{-1/2}$ \citep{fournier2015rate}. At the pilot
size $n=200$ its 95th percentile ($\approx 0.22$) is much larger than a
fixed threshold of 0.05, so naive thresholding ``confirms'' pure noise.
For this reason all tests in the audit are calibrated against per-state
nulls at $n = 2{,}000$.}
\label{fig:noise}
\end{figure}

\textbf{P1: RNG cloning.} Restart-based evaluation usually takes
snapshots of environments with \texttt{deepcopy}, and this also clones
the internal random generator of the environment. As a result every
``random'' rollout replays the same future, and the \emph{ground truth}
collapses to a point mass. Our first audit was built in this way, and it
concluded that all claimed risk was an artifact. We noticed the problem
only because a game with random spawns was behaving deterministically.
The failure gives no error, the wrong conclusion looks plausible, and it
can easily pass a code review.

\textbf{P2: the empirical Wasserstein noise floor.} For two $n$-samples
from the \emph{same} distribution with spread $\sigma$, the empirical
$\excess$ concentrates around $c\,\sigma n^{-1/2} > 0$ and not around
zero \citep{fournier2015rate}. Figure~\ref{fig:noise} shows this floor
on real audit data. At our pilot size of $n = 200$ the mean floor is
${\approx}0.07$ and its 95th percentile is ${\approx}0.22$, which is
more than most true effect sizes. In our second audit, a fixed small
threshold ``confirmed'' two-thirds of the claims, and all of these were
noise. Hence any validation of distributional claims has to be
calibrated against a per-state null at the sample size actually used,
which is why we use the permutation and bootstrap tests described above.

\section{Experiments}
\label{sec:experiments}

\begin{figure*}[t]
\centering
\includegraphics[width=0.44\textwidth]{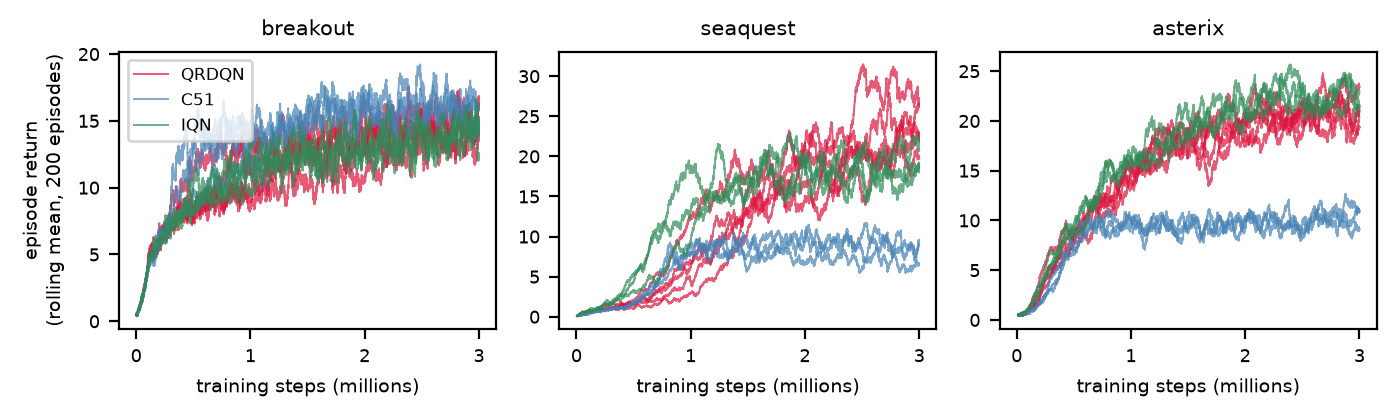}
\caption{Training curves for all 33 MinAtar agents (rolling mean over
200 episodes). Final scores (mean of the last 100 episodes, averaged
over seeds) are QR-DQN 14.6 / 25.2 / 22.1, C51 15.8 / 9.2 / 10.0 and IQN
13.4 / 17.7 / 22.3 on Breakout / Seaquest / Asterix, which is in line
with published MinAtar baselines \citep{young2019minatar}. C51 does
worse than QR-DQN on two games. Since the audit is within-agent (each
head is tested against its own claims), the audit conclusions do not
depend on the scores being equal across algorithms.}
\label{fig:training}
\end{figure*}

\textbf{Setup.} We train QR-DQN (51 quantiles, 3M steps, five seeds),
C51 and IQN (three seeds each) on MinAtar \citep{young2019minatar}
Breakout, Seaquest and Asterix, with a shared recipe given in the
supplement and MinAtar defaults including sticky actions ($p{=}0.1$).
The training curves are shown in Figure~\ref{fig:training}. In addition,
we audit intermediate Breakout checkpoints (0.5M/1M/2M steps) and the
\emph{pretrained} SB3-zoo QR-DQN agents (10M ALE frames, 200 quantiles)
on full-scale ALE Breakout and Seaquest ($n{=}1{,}000$ rollouts, sticky
evaluation with $p{=}0.25$). Together with the positive controls below,
this comes to 2{,}450 audited states and 9.6M rollouts. C51 claims are
quantized to its atom grid (bin width $0.8$), so comparisons across
algorithms are restricted to claims above this width. The whole sweep
cost less than \$9 of commodity CPU time. All code, checkpoints and raw
returns are released.

\subsection{The Strongest Risk Claims Are Mostly False}

\begin{table}[t]
\centering
\footnotesize
\setlength{\tabcolsep}{2.6pt}
\begin{tabular}{lccc}
\toprule
QR-DQN & Breakout & Seaquest & Asterix \\
\midrule
Refuted, top 2\% of ranking & \textbf{84\%} {\tiny[74,91]} & \textbf{84\%} {\tiny[74,91]} & \textbf{66\%} {\tiny[54,76]} \\
Refuted, rest of ranking & 20\% {\tiny[15,27]} & 15\% {\tiny[10,21]} & 5\% {\tiny[2,9]} \\
Refuted, top 2\%, low-alias & 80\% {\tiny[66,90]} & 100\%$^\dagger$ & 91\% {\tiny[62,98]} \\
Confirmed (FDR 0.1), pooled & 0/245 & 0/245 & 0/245 \\
corr(learned, true excess) & 0.02 & $-$0.01 & $-$0.04 \\
\bottomrule
\end{tabular}
\caption{Main results over 5 seeds per game (245 audited states per game,
Wilson 95\% intervals). $^\dagger$Only one low-alias state in this
stratum. The corr rows are per-seed means.}
\label{tab:main}
\end{table}

\begin{figure}[t]
\centering
\includegraphics[width=0.8\columnwidth]{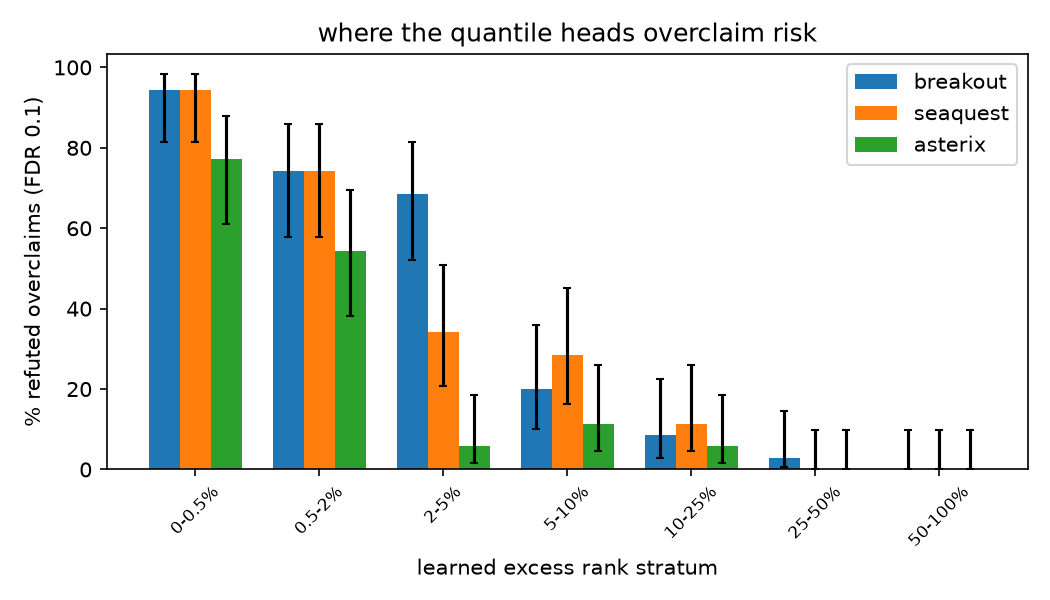}
\caption{Refutation rate by rank stratum (QR-DQN, pooled over 5 seeds,
Wilson 95\% intervals, FDR 0.1). The most confident risk claims of the
model (left) are the least true. Below the top decile of the ranking the
refutations mostly disappear, because the claims there are too small to
be provably wrong. A truth-blind head with the same claim magnitudes
gives a similar profile (see text).}
\label{fig:refutation}
\end{figure}

In the top 2\% of the agent's own $\excess$ ranking, 66--84\% of the
claimed risk trade-offs are refuted at 95\% confidence
(Table~\ref{tab:main}). For these states the learned excess is larger
than anything the environment-grounded distributions could plausibly
produce, even though the bias of the estimator works in favour of the
agent. Aliasing does not explain this rate. Among low-alias states,
where the observation more or less fixes the outcome distribution, the
refutation rate is the same or higher. Refutation decreases
monotonically as we go down the ranking (Fig.~\ref{fig:refutation}),
from ${\approx}94\%$ in the top half-percent of Breakout and Seaquest to
nearly zero below the top quartile.

This gradient has to be interpreted carefully. Small claims cannot be
falsified by construction, since refutation needs the claim to be larger
than the MC bootstrap percentile, and so some decline is expected from
this asymmetry of the test alone. A truth-blind null head allows an
exact comparison. We permute the learned claims of each run across its
audited states, which keeps the magnitudes and rank strata but
randomizes the placement with respect to ground truth, and then re-run
the same pipeline. This gives a top-strata refutation of $81\%$
($[75, 86]$ over $1{,}000$ permutations), compared with $79\%$ observed
($165/210$, $p = 0.22$). So the heads do no better than truth-blind
chance in where they put their strongest claims.

\subsection{Reported Uncertainty Is Uncorrelated with Real Uncertainty}

\begin{figure*}[t]
\centering
\includegraphics[width=0.46\textwidth]{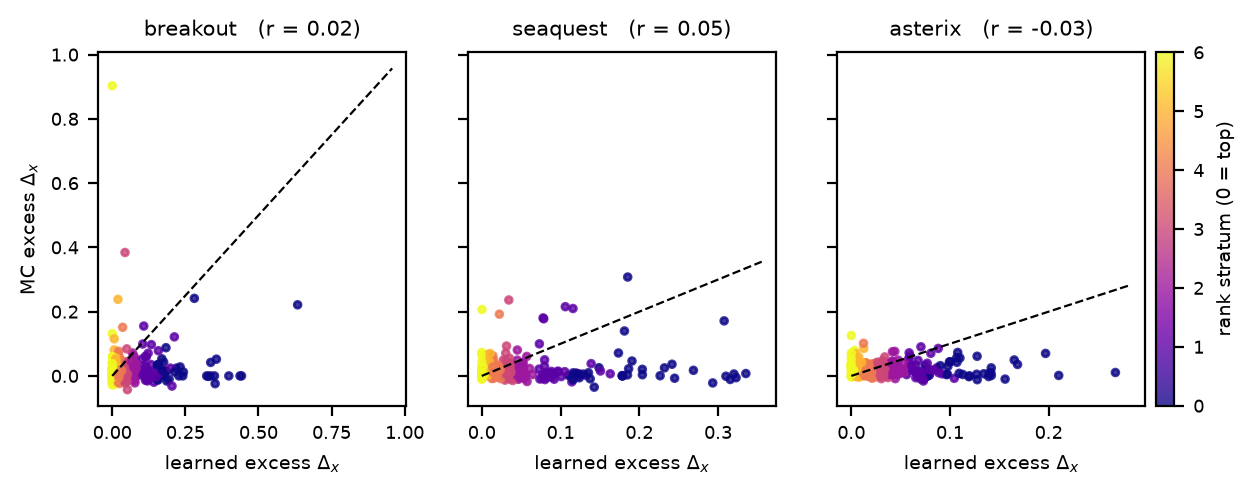}
\caption{Learned excess vs.\ environment-grounded (MC) excess for all
735 audited QR-DQN states (color = rank stratum, 0 = top of ranking;
dashed = identity; $r$ = pooled Pearson correlation). If the heads were
calibrated even in a noisy way, the top-stratum (dark) points would lie
close to the diagonal. Instead they lie flat along the floor.}
\label{fig:scatter}
\end{figure*}

Figure~\ref{fig:scatter} plots the learned excess against the true
excess for all audited states. The pooled correlation is $0.02$ /
$0.05$ / $-0.03$ for the three games (per-seed means $0.02$ / $-0.01$ /
$-0.04$, and no seed is significantly positive). So the ranking which a
practitioner would use to locate ``the states where risk matters''
actually shows where the representation is least trustworthy. Further,
no claim can be confirmed. Our tests have the power to detect true
trade-offs of magnitude ${\gtrsim}0.07$ (the median per-state null), and
yet the number of pooled FDR-corrected confirmations is \emph{zero} in
all three games. Uncorrected per-state detections at $p < 0.05$ stay
around the false-positive rate ($\approx 5\%$). There are 9 raw per-seed
detections among 735 QR-DQN states and 7 among 441 C51 states, and none
of them survive Benjamini--Hochberg pooling. Under greedy play, the
aleatoric noise of MinAtar does not create action-conditional dominance
violations of the size that the heads commonly claim. It appears that
the heads are reporting trade-offs which do not exist at all, and that
this is not a case of real trade-offs being exaggerated.

\subsection{The Audit Passes Its Positive Control}
\label{sec:positive}

\begin{figure*}[t]
\centering
\includegraphics[width=0.72\textwidth]{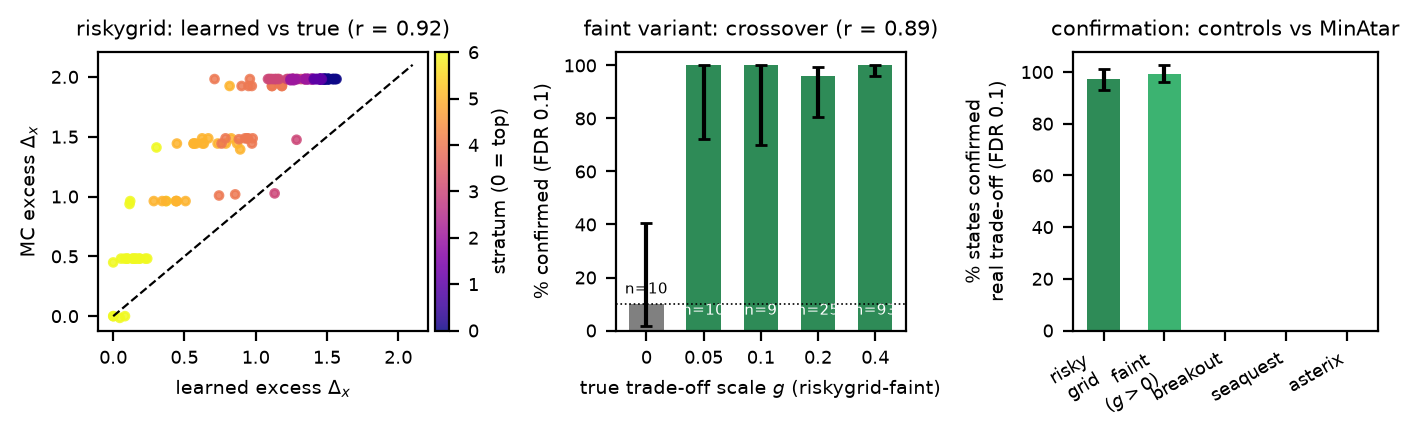}
\caption{Positive controls. Left: on \textsc{riskygrid}, where trade-offs
exist by construction, the learned ranking follows ground truth (pooled
$r = 0.92$, per-seed $0.96/0.92/0.89$). The head compresses the
magnitudes but orders them correctly (compare with the flat clouds of
Fig.~\ref{fig:scatter}). Center: the \emph{faint} variant, with
trade-offs of scale $g \in \{0, 0.05, \dots, 0.4\}$ around the detection
floor of the audit. Here 96--100\% of $g>0$ states are confirmed and
$g{=}0$ states stay at the FDR budget. Right: FDR-corrected confirmation
rate for the controls and for the three MinAtar games (QR-DQN, Wilson
95\% intervals). All the MinAtar bars are at zero.}
\label{fig:positive}
\end{figure*}

A result of zero confirmations naturally raises the objection that the
audit may be broken and the model may be fine. To check this we built
an environment with a known answer key. \textsc{riskygrid} is a
ten-step corridor. In each episode, per-column gamble scales $g(x) \in
\{0, 0.5, \dots, 2.0\}$ are drawn and rendered into the observation. At
each step the agent chooses either \textsc{safe} (reward $0.5$,
deterministic) or \textsc{gamble} (reward $0.5 \pm g(x)$ with a fair
coin). The means are identical, so every column with $g(x) > 0$ is a
pure risk trade-off of known magnitude and columns with $g(x)=0$ have no
trade-off. The observation contains the full $g$-map, so there is no
aliasing by construction. We train QR-DQN (three seeds, 500k steps) and
run exactly the same audit.

On this control the verdicts are reversed (Fig.~\ref{fig:positive}). We
get $143/147 = 97\%$ $[93, 99]$ pooled FDR-corrected confirmations, and
the misses are all $g{=}0$ states. There are \emph{zero} false
refutations in the top strata ($0/42$, and $4/105$ below, which is
consistent with the FDR budget). The learned--true correlation is $0.92$,
whereas $|r| \le 0.11$ everywhere on MinAtar, and the CVaR choice of the
head picks the truly safer action in $99$--$100\%$ of 143 decidable
states with realized regret of ${\approx}0$. We also made a \emph{faint}
variant in which the gambles are scaled down to $g \in \{0, 0.05, 0.1,
0.2, 0.4\}$. These values lie around both the detection floor of the
audit and the scale of the artifact spread on MinAtar. The audit
confirms 96--100\% of $g > 0$ states at every level including $g = 0.05$
(Fig.~\ref{fig:positive}, center), keeps $g = 0$ states at the FDR
budget (1/10), and the ranking still follows the truth with pooled
$r = 0.89$. The same architecture and the same audit which report
spurious risk on MinAtar report almost perfect risk here. Therefore the
MinAtar result cannot be explained by low power of the audit or by
small effect sizes.

\subsection{Does Acting on the Learned Uncertainty Matter?}
\label{sec:cvar}

\begin{table}[t]
\centering
\footnotesize
\setlength{\tabcolsep}{3.4pt}
\begin{tabular}{llcccc}
\toprule
 & & \multicolumn{2}{c}{safer pick (\%)} & \multicolumn{2}{c}{CVaR regret} \\
 \cmidrule(lr){3-4} \cmidrule(lr){5-6}
 & & head & mean & head & mean \\
\midrule
\multirow{3}{*}{\shortstack[l]{QR-DQN\\($\alpha{=}0.1$)}}
 & Breakout (n=27) & \textbf{81} & 44 & \textbf{0.20} & 1.36 \\
 & Seaquest (n=35) & 34 & \textbf{63} & 0.65 & \textbf{0.30} \\
 & Asterix (n=27)  & 52 & 48 & 0.20 & 0.26 \\
\midrule
\multirow{3}{*}{\shortstack[l]{C51\\($\alpha{=}0.1$)}}
 & Breakout (n=17) & 65 & 71 & 0.45 & \textbf{0.10} \\
 & Seaquest (n=18) & 28 & \textbf{56} & 0.67 & \textbf{0.20} \\
 & Asterix (n=14)  & 36 & 57 & 0.48 & \textbf{0.24} \\
\bottomrule
\end{tabular}
\caption{Decision-level evaluation at the most-flagged (top strata)
decidable states of each head. We report the \% of states where each
chooser picks the truly safer action, and the mean realized CVaR regret
with respect to the ground-truth-optimal choice. \textbf{Bold} marks a
clear winner. The mean-greedy column ignores risk completely. The IQN
rows and $\alpha{=}0.25$ (quoted in the text) show the same pattern.
Per-state data is released.}
\label{tab:cvar}
\end{table}

For anybody who uses learned uncertainty, the practical test is whether
it leads to better decisions. When the model says that $A$ is safer than
$B$, we can check with rollouts whether $A$ really has the better risk
profile in the environment. Using the same rollouts as before, we
compare three ways of choosing the safer action at level $\alpha$ on
\emph{decidable} states (states where the true CVaR gap is significant
at 95\% by bootstrap). These are the CVaR ranking of the head, the
mean-greedy baseline, and ground truth (Table~\ref{tab:cvar}).

The outcome depends strongly on the environment. In Breakout the QR-DQN
head is genuinely informative. In the top strata at $\alpha{=}0.1$ it
chooses the safer action 81\% of the time against 44\% for mean-greedy,
with a regret of $0.20$ against $1.36$. In Seaquest it is
\emph{anti-predictive}. At the states that it flags as most
risk-critical, its CVaR choice is truly safer only 24--34\% of the time,
which is below chance (one-sided binomial $p = 8\times 10^{-4}$ at
$\alpha{=}0.25$ and $p = 0.045$ at $\alpha{=}0.1$), and the regret is
roughly three times that of ignoring risk altogether. In Asterix the
head is no better than a coin flip. C51 does worse, and is beaten by
mean-greedy in all three games (28--68\% vs.\ 48--71\%, with
2--4.5$\times$ the regret). IQN shows the same regimes as QR-DQN. It is
helpful in Breakout (79\% vs.\ 42\%), at chance level in Seaquest, and
inverted in Asterix (47\% vs.\ 73\%). A practitioner has no way of
knowing which regime applies. Advice that is sometimes helpful and
sometimes inverted is harder to use than advice that is uniformly
uninformative, and for none of the three algorithms could we find a
rule, independent of the game, that tells when the risk advice of the
model can be trusted.

\subsection{Specimens: What Spurious Uncertainty Looks Like}

\begin{figure}[t]
\centering
\includegraphics[width=0.6\columnwidth]{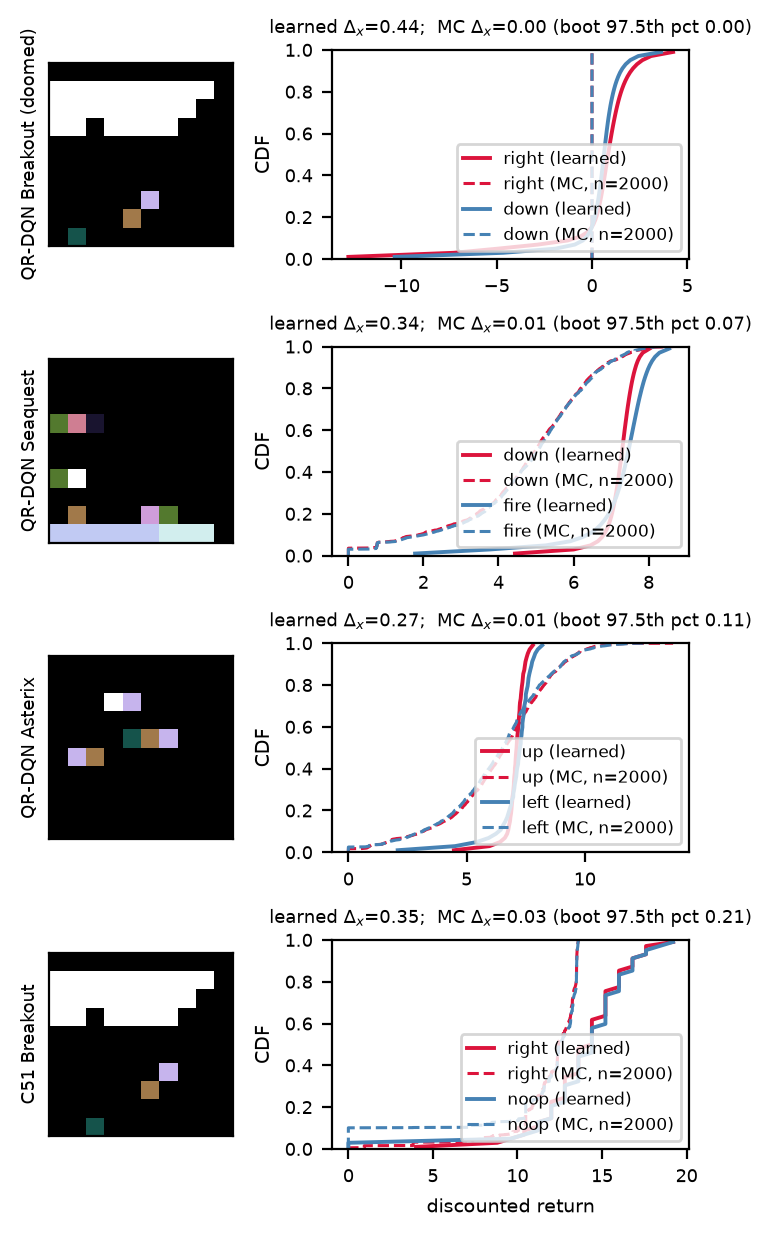}
\caption{Audited specimens. Each row shows the environment snapshot
(left) and the return CDFs of both actions, learned (solid) vs.\
2{,}000-rollout MC ground truth (dashed). Top: a \emph{doomed} Breakout
state. Every future gives the same return (dashed step), but the head
reports crossing distributions with excess $0.44$. Middle: top-stratum
Seaquest and Asterix states where the true distributions coincide and
the learned ones cross. Bottom: the same signature in C51.}
\label{fig:gallery}
\end{figure}

The clearest specimens (Fig.~\ref{fig:gallery}) are Breakout states in
which the ball has been lost and cannot be recovered. All $2{,}000$
futures give exactly the same return (the bootstrap 97.5th percentile of
the MC excess is $0.00$), but the head reports wide, crossing
distributions with learned excess as high as $0.44$. Pooled over seeds,
$14/70 = 20\%$ of top-stratum Breakout states are of this doomed type.
At these states there is no uncertainty at all in the environment, and
still the model reports a risk trade-off. The Seaquest and Asterix
specimens show the more common failure. Here the true distributions of
the two actions cannot be told apart, while the head claims a
decision-relevant asymmetry which exists only in the network.

\subsection{Replication: C51, IQN, and a Pretrained Atari Agent}
\label{sec:c51}

\begin{table}[t]
\centering
\footnotesize
\setlength{\tabcolsep}{2.3pt}
\begin{tabular}{llccc}
\toprule
 & & Breakout & Seaquest & Asterix \\
\midrule
\multirow{4}{*}{C51}
 & Refuted, top 2\% & \textbf{81\%} {\tiny[67,90]} & \textbf{95\%} {\tiny[84,99]} & \textbf{57\%} {\tiny[42,71]} \\
 & Refuted, rest & 10\% {\tiny[5,17]} & 16\% {\tiny[10,24]} & 5\% {\tiny[2,11]} \\
 & Confirmed, pooled & 0/147 & 0/147 & 0/147 \\
 & corr(learned, true) & 0.02 & 0.05 & $-$0.11 \\
\midrule
\multirow{4}{*}{IQN}
 & Refuted, top 2\% & \textbf{64\%} {\tiny[49,77]} & \textbf{79\%} {\tiny[64,88]} & \textbf{40\%} {\tiny[27,56]} \\
 & Refuted, rest & 7\% {\tiny[3,13]} & 7\% {\tiny[3,13]} & 8\% {\tiny[4,14]} \\
 & Confirmed, pooled & 0/147 & 0/147 & 0/147 \\
 & corr(learned, true) & 0.08 & $-$0.11 & $-$0.08 \\
\midrule
\multirow{4}{*}{\shortstack[l]{ALE\\(zoo)}}
 & Refuted, top 2\% & \textbf{100\%} {\tiny[78,100]} & 14\% {\tiny[4,40]} & n/a \\
 & Refuted, rest & 0\% {\tiny[0,10]} & 3\% {\tiny[1,15]} & n/a \\
 & Confirmed, pooled & 0/49 & 0/49 & n/a \\
 & corr(learned, true) & $-$0.22 & $-$0.25 & n/a \\
\bottomrule
\end{tabular}
\caption{C51 and IQN audits with 3 seeds per game (147 audited states
per game per algorithm, Wilson 95\% intervals, FDR 0.1, corr rows are
per-seed means), and full-Atari audits of the single pretrained SB3-zoo
QR-DQN for each game ($n{=}1{,}000$ rollouts, sticky-action
evaluation). The QR-DQN signature of Table~\ref{tab:main} is seen again
under a categorical projection, under an implicit-quantile projection,
and for a professionally trained 10M-frame agent on the full ALE.}
\label{tab:c51}
\end{table}

\begin{figure}[t]
\centering
\includegraphics[width=0.72\columnwidth]{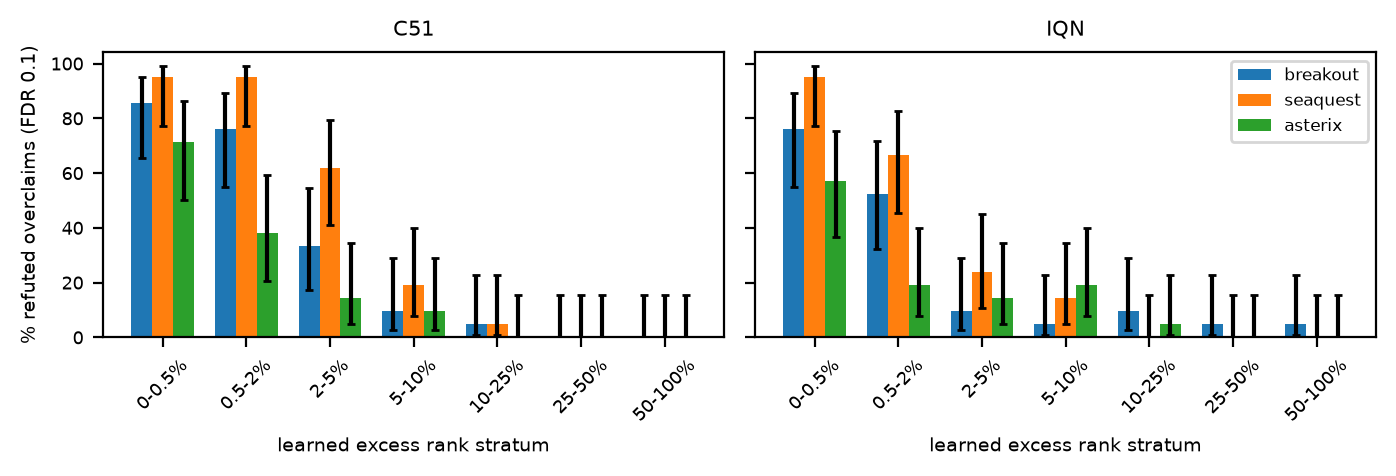}
\caption{Refutation rate by rank stratum for C51 (left) and IQN (right),
pooled over 3 seeds (Wilson 95\% intervals, FDR 0.1). The decline from
the top of the ranking is monotone, as in Fig.~\ref{fig:refutation}.}
\label{fig:refutation-c51}
\end{figure}

When we repeat the full audit for C51 and IQN (Table~\ref{tab:c51},
Fig.~\ref{fig:refutation-c51}), we find the same signature under two
more projections. In all 18 runs, 40--95\% of top-strata claims are
refuted against a background of 5--16\%, there are zero pooled
confirmations among 882 states, and there is no consistent positive
correlation. The categorical projection of C51 does not have any of the
fixed-point pathologies of QTD \citep{rowland2018analysis,rowland2023qtd},
and IQN does away with the fixed $\tau$ grid entirely. The overclaiming
therefore seems to come from the ingredient that all three share, which
is bootstrapped distributional targets from a changing policy under
function approximation, and the choice of representation does not seem
to matter much.

The same signature is also seen at full-Atari scale, with one caveat.
The zoo agents were trained \emph{without} sticky actions, so by
construction their claimed spread cannot come from environment
stochasticity. This audit therefore measures the size and placement of
the artifact, and should not be taken as an independent replication
(Table~\ref{tab:c51}, bottom). For the pretrained SB3-zoo QR-DQN on
Breakout (raw score ${\approx}400$, which matches its published
performance), \emph{all} 14 top-strata claims are refuted and none
elsewhere. There are zero confirmations among 49 states and the
correlation is negative, with a mean top-strata claim of $0.184$ against
a mean ground-truth excess of $0.000$. The zoo Seaquest head sits at the
null floor (mean $0.021$ vs.\ $0.03$), so only $2/14$ claims can be
refuted. Even so, nothing can be confirmed and the correlation is again
negative. On MinAtar, 32 of the 33 runs refute at least $6/14$
top-strata claims against a background of 5--20\% (per-run counts are
released).

\subsection{Structural, Not a Symptom of Weak Models}
\label{sec:structural}

\begin{figure}[t]
\centering
\includegraphics[width=0.62\columnwidth]{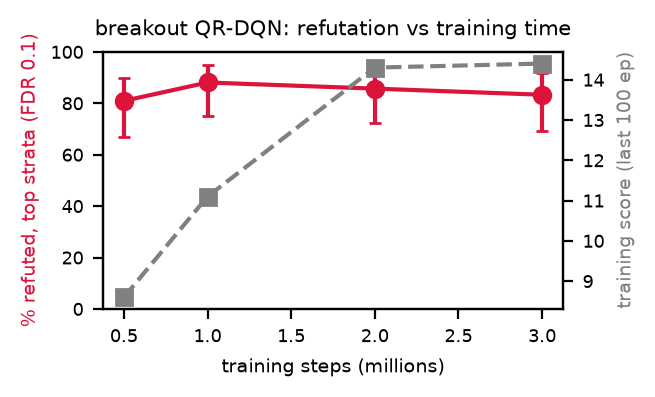}
\caption{Full audits of intermediate checkpoints (Breakout QR-DQN, 3
seeds pooled, 42 top-strata states per point). Top-strata refutation is
$81/88/86/83\%$ at $0.5/1/2/3$M steps, while the training score goes up
from 8.6 to 14.4. There are no confirmations at any stage.}
\label{fig:traintime}
\end{figure}

Three probes connect the artifact to the learning process and not to
how well the agent plays. Firstly, \emph{training time.} Audits of
intermediate checkpoints (Fig.~\ref{fig:traintime}) show that the
overclaiming is fully formed by 500k steps and stays the same up to 3M
steps, even though the score nearly doubles in this period. The claimed
magnitude in the top 2\% actually \emph{grows} over training, in 15 out
of 15 runs. Secondly, \emph{performance.} Across all 33 runs, score and
top-strata refutation are uncorrelated (Spearman $-0.07$), so better
agents do not make better claims. Thirdly, \emph{seeds.} Under the four
sibling seeds of the same game, the siblings reproduce a median of only
$3.8\%$ of the magnitude of a top-stratum claim, and the cross-seed
median remains uncorrelated with truth ($|r| \le 0.07$). Each seed thus
produces its own spurious claims. From these probes we conclude that the
learned spread comes from the training process (bootstrapped targets
from a changing policy, the projection step, optimization noise) and
that environment stochasticity is not its source. This is a per-state,
behavioral instance of the mixing that \citet{clements2019estimating}
identified at the level of the architecture. The spurious claims gather
at the top of the ranking because of selection. Ranking by $\excess$
picks out the largest noise values, which is a kind of winner's curse,
and so face-value readings are least reliable at the states most likely
to be looked at.

\subsection{Repairs Do Not Fix It}
\label{sec:repairs}

We tried three repairs which one would expect to help from the
literature, and none of them gives usable risk claims (3 seeds per
game). \emph{Risk-sensitive training.} A QR-DQN trained to be
$\cvar_{0.25}$-greedy (it selects and bootstraps from the CVaR of its
own quantiles, and is audited under the matching policy) shows the same
signature as before. Top-strata refutation is $88$--$98\%$, pooled
confirmations are zero in all three games, and pooled correlation is
$0.06$/$0.04$/$-0.02$, although the scores are healthy (Breakout $16.7$,
Asterix $18.4$). Optimizing for the tail does not make the claims about
the shape of the tail true. \emph{Ensembling.} A five-member ensemble on
a shared buffer, audited at the average of its quantile functions (the
$\Wone$ barycenter), \emph{reduces} the artifact. Top-strata refutation
comes down from $95$/$100$/$95\%$ for a single member to $81$/$98$/$57\%$,
and $8$ Breakout confirmations are recovered. Seaquest and Asterix,
however, still have zero confirmations, and the pooled correlation
remains close to zero ($0.14$/$0.12$/$-0.11$). Averaging helps to some
extent, but the top claims are still not calibrated.

\textbf{Miscalibrated or uninformative?} We fit the best monotone
recalibration (an isotonic map from learned to true excess) on a
calibration split and apply it on a held-out split. On the positive
controls this \emph{rescues} the head, giving zero refutations,
$98$--$99\%$ of genuine trade-offs asserted, and correlation of
${\approx}0.9$. On MinAtar, on the other hand, the only recalibration
that reduces refutation shrinks the top claims to $12$--$22\%$ of their
magnitude and asserts $0\%$ of any real trade-off. It passes the audit
only by making no claims at all. We take this to mean that the head is
\emph{uninformative} about the true risk. The information is missing,
and this cannot be fixed by rescaling. (A distillation control in the
supplement confirms that the architecture is able to fit the true
distributions when it is supervised on them.) It is also not possible
to flag the unsupported claims using the model alone. A leak-free probe
on 13 head-shape features does no better than the trivial
claim-magnitude baseline (AUROC $0.92$ in both cases). Passing a
calibration check is thus not sufficient, since a head can be
recalibrated to pass while having no information that helps in making
a decision.

\section{Discussion}
\label{sec:discussion}

\textbf{Decision-relevant uncertainty.} The main observation of this
work is that a model can produce a valid predictive distribution whose
spread is still unreliable about the risks that matter for decisions.
Evaluations that score the distribution on average are unlikely to
catch this. The errors are concentrated at the states where the model
reports the largest risk differences, and these are the very states on
which a risk-sensitive decision rule depends. We therefore suggest that
learned uncertainty be evaluated at the decisions where it is used. One
first finds the states where the uncertainty of the model would change
a decision, and then checks whether the implied difference in risk is
seen in environment-grounded outcomes. The excess gap is one way of
doing this, and per-decision CVaR agreement (Table~\ref{tab:cvar}) is
another.

\textbf{Implications for learned world models.} If a world model is
evaluated only by likelihood, calibration or distributional similarity,
it may not be clear whether its uncertainty is useful to a planner. Our
recalibration result shows that a distribution can pass a calibration
check and still be uninformative. Our ensembling result shows that
averaging does not necessarily bring back the risk structure on which
decisions depend. The audit suggests one more check, which is to find
the decisions that depend on the uncertainty of the model and to test
whether the predicted risk differences hold up against real outcomes.
In a robotic setting, the corresponding procedure would compare the
future-state or return distributions predicted by the model with
repeated interactions with the environment, starting from a simulator
reset or a controlled physical reset. The rollout budget would be spent
mainly on states where the model predicts clearly different levels of
risk. The stratified sampling, per-state nulls and FDR control of
\S\ref{sec:method}, as well as the two pitfalls of
\S\ref{sec:pitfalls}, would apply as they are. We present this as a
complementary way of evaluating such models. It is to be noted that our
experiments do not establish how robotic world models behave.

\textbf{Implications for distributional RL.} (i) Interpretability claims
that are based on visualized return distributions need to be validated
against ground truth, as was the case for saliency maps
\citep{atrey2020exploratory}. (ii) Risk-sensitive policies derived from
raw quantile heads \citep{keramati2020cvar,lim2022risksensitive} end up
acting on spurious risk at the critical states. (iii) Repairs which
build uncertainty estimates around the raw head
\citep{eriksson2022sentinel,zanger2024dpe,stutts2024ceqr,gan2025conformal}
now have a measurement against which they can be evaluated.

\textbf{Limitations and scope.} Our experiments are not robotic, and we
audit distributional value functions and not learned dynamics or world
models. The connection made above is therefore methodological and not
empirical. The audit depends on environment-grounded distributions,
which we can obtain here only because the simulator can be restarted
from snapshots and sampled thousands of times per state. Physical
systems allow this only approximately and with far fewer samples, and
the noise-floor analysis (P2) shows that fewer samples come at a real
cost. The return is a scalar summary. To audit high-dimensional
predictive distributions (over future observations or latent states),
one has to choose projections of them that are relevant to decisions,
and extending the audit to partially observed, high-dimensional robotic
world models is left for future work. Within distributional RL, our
claims cover QR-DQN, C51 and IQN under greedy and $\cvar$-greedy
evaluation, ensembles, and one pretrained QR-DQN per ALE game. Other
calibrated variants \citep{stutts2024ceqr,gan2025conformal} may behave
differently, and an ALE agent trained with sticky actions is still the
missing test at scale. For regimes where faint risk is entangled with
the dynamics, we do not have an answer key.

\section{Conclusion}

In this paper we asked when the uncertainty reported by a learned model
can be trusted for decisions. We studied this question in a setting
where the answer can be checked, namely distributional RL agents, whose
learned return distributions we compared state by state with
environment-grounded ones. For the agents that we tested, the
uncertainty mostly cannot be trusted. On MinAtar, $40$--$95\%$ of the
strongest claimed risk trade-offs are provably false, almost none can be
confirmed, and the placement of the most confident claims is
indistinguishable from truth-blind. The effect is structural, is also
seen for a pretrained near-state-of-the-art Atari agent, and remains
after risk-sensitive training, ensembling and recalibration. Acting on
these claims gives results from helpful to worse than chance, and the
practitioner has no observable signal to know which regime they are in.
On positive controls with trade-offs of known magnitude, the same audit
recovers the real risk.

More generally, a predictive distribution makes claims about the world,
and its uncertainty should be checked against realized outcomes at the
decisions where it is used, along with aggregate measures of fit. We
release the audit toolkit and propose a simple standard for
uncertainty-aware learned models, world models included. A model that
passes the audit should be able to confirm its top claims.

\setlength{\bibsep}{1.5pt plus 0.3pt}
\bibliography{references}

\end{document}